%% file: iccc.tex
\documentclass[letterpaper]{article}
\usepackage{iccc}
\usepackage{times}
\usepackage{helvet}
\usepackage{courier}
\usepackage{xcolor}
\usepackage{url}
\usepackage[scaled=0.90]{inconsolata}

\usepackage{graphicx}
\usepackage{amsmath,amssymb}
\usepackage{color}
\usepackage{epsfig}
\usepackage{pdfpages}
\usepackage{caption}
\usepackage{subcaption}

\usepackage{multirow}
\usepackage{rotating}
\usepackage{booktabs}
\usepackage{algpseudocode}
\usepackage{algorithm}
\usepackage{etoolbox}
\usepackage{color,soul}
\usepackage{hyperref}

\usepackage{xcolor}
\definecolor{orange}{rgb}{1,0.5,0}
\definecolor{lightsalmonpink}{rgb}{1.0, 0.6, 0.6}
\definecolor{verylightsalmonpink}{rgb}{0.966, 0.805, 0.797}
\definecolor{lightblue}{rgb}{0.862, 0.906, 0.984}
\definecolor{lightyellow}{rgb}{1.0, 0.945, 0.797}
\definecolor{lightgreen}{rgb}{0.835, 0.91, 0.828}
\definecolor{lightpurple}{rgb}{0.879, 0.832, 0.902}
\definecolor{champagne}{rgb}{0.97, 0.91, 0.81}
\definecolor{bubbles}{rgb}{0.91, 1.0, 1.0}
\definecolor{belize}{RGB}{41, 128, 185}

\newcommand{\state}[0]{\setlength{\fboxsep}{0pt}\colorbox{champagne}{\textsc{ST}}}
\newcommand{\action}[0]{\colorbox{bubbles}{\textsc{AC}}}
\newcommand{\stateplusaction}[0]{\colorbox{lightyellow}{\textsc{SA}}}
\newcommand{\syncseq}[0]{\colorbox{lightblue}{\textsc{SS}}}
\newcommand{\unsyncseq}[0]{\colorbox{lightgreen}{\textsc{US}}}
\newcommand{\syncran}[0]{\colorbox{lightpurple}{\textsc{SR}}}
\newcommand{\unsyncran}[0]{\colorbox{verylightsalmonpink}{\textsc{UR}}}

\newtoggle{arxiv}

\title{Feel The Music: Automatically Generating A Dance For An Input Song}

\iftoggle{arxiv}{
\author{Jane Smith$^{1,2}$\\
$^{1}$Georgia Tech\\
$^{2}$Allen Institute for Artificial Intelligence\\
$^{3}$Facebook AI Research \\
smith@jane.com \quad smith@jane.com\\
}
} {
\author{
    Purva Tendulkar$^{1}$ \hspace{1.5pc}
    Abhishek Das$^{1\rightarrow3}$ \hspace{1.5pc}
	Aniruddha Kembhavi$^{2}$ \hspace{1.5pc}
    Devi Parikh$^{1,3}$ \\
    $^1$Georgia Institute of Technology,~~
	$^2$Allen Institute for Artificial Intelligence,~~
	$^3$Facebook AI Research\\
	$^1${\tt \{purva, parikh\}@gatech.edu},~~
	$^2${\tt anik@allenai.org},~~
	$^3${\tt abhshkdz@fb.com}
}
}

\begin{document}
\maketitle
\input{sections/abstract}
\input{sections/intro}
\input{sections/related_work}
\input{sections/dataset}
\input{sections/approach}
\input{sections/evaluation}
\input{sections/discussion}

\bibliographystyle{iccc}
\bibliography{iccc}

\end{document}

%% file: sections/abstract.tex
\begin{abstract}
\begin{quote}

We present a general computational approach that enables a machine to generate a dance  for any input music. 
We encode intuitive, flexible heuristics for what a `good' dance is: the structure of the dance should align with the structure of the music. This flexibility allows the agent to discover creative dances. 
Human studies show that participants find our dances to be more creative and inspiring compared to meaningful baselines. 
We also evaluate how perception of creativity  changes based on different presentations of the dance. 
Our code is available at \href{https://github.com/purvaten/feel-the-music}{{\tt github.com/purvaten/feel-the-music}}.
\end{quote}
\end{abstract}

%% file: sections/intro.tex
\section{Introduction}
\label{sec:intro}

Dance is ubiquitous human behavior, dating back to at least $20,000$ years ago~\cite{appenzeller1998evolution}, and embodies human self-expression and creativity. At an `algorithmic' level of abstraction, dance involves body movements organized into spatial patterns synchronized with temporal patterns in musical rhythms. Yet our understanding of how humans represent music and how these representations interact with body movements is limited~\cite{brown2006neural}, and computational approaches to it under-explored.

We focus on automatically generating creative dances for a variety of music. 
Systems that can automatically recommend and evaluate dances for a given input song can aid choreographers in creating compelling dance routines, inspire amateurs by suggesting creative moves, and propose modifications to improve dances humans come up with.
Dancing can also be an entertainment feature in household robots, much
like the delightful ability of today's voice assistants to tell jokes or
sing nursery rhymes to kids!

Automatically generating dance is challenging for several reasons. 
First, like other art forms, dance is subjective, which makes it hard to computationally model and evaluate. 
Second, generating dance routines involves synchronization between past, present and future movements whilst also synchronizing these movements with music. 
And finally, compelling dance recommendations should not just align movements to music, they should ensure these are enjoyable, creative, and appropriate to the music genre.

\begin{figure}[t]
\centering
\includegraphics[width=0.93\columnwidth]{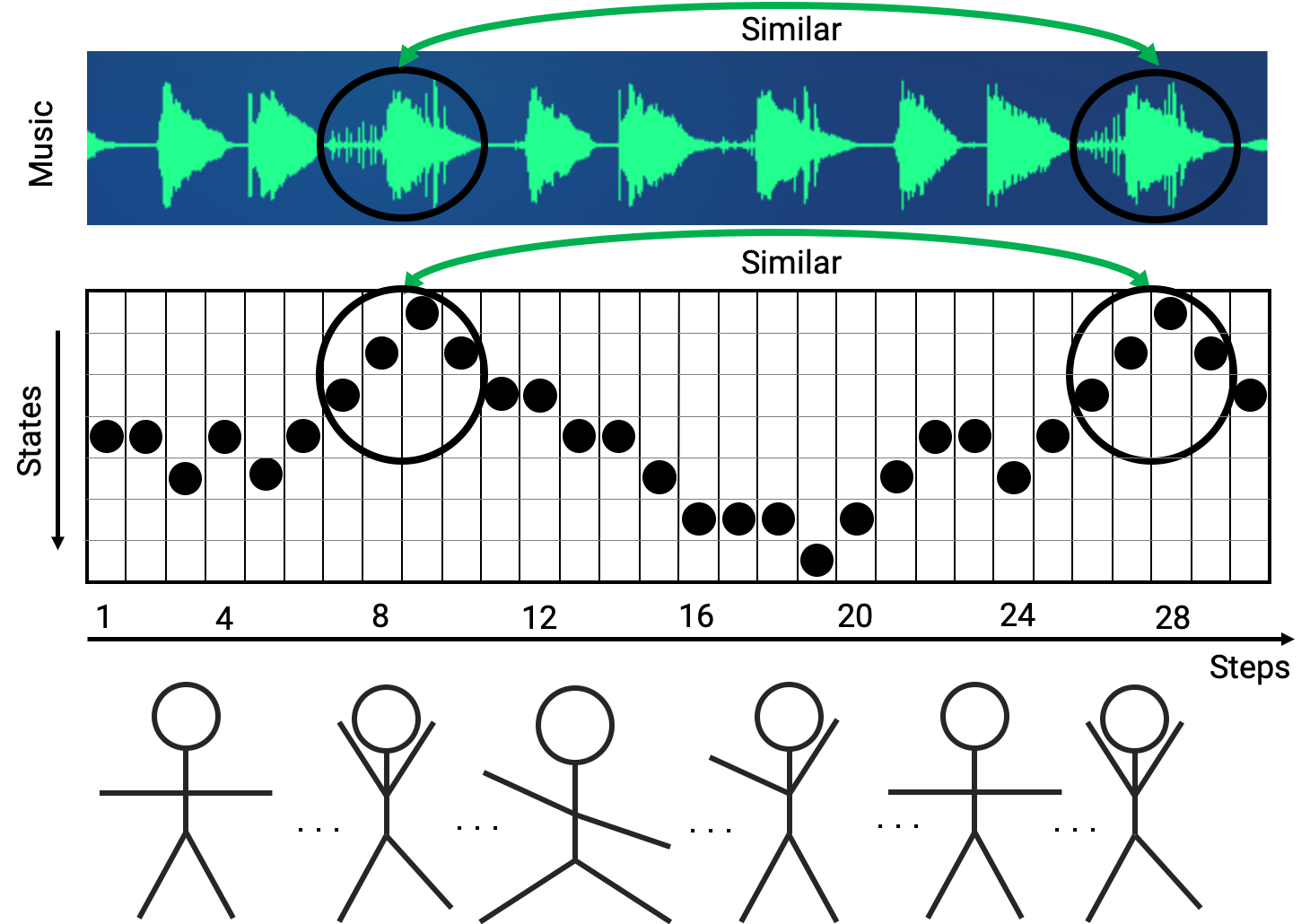}
\vspace{-4pt}
\caption{Given input music (top), we generate an aligned dance choreography as a sequence of discrete states (middle) which can map to a variety of visualizations (e.g., humanoid stick-figure pose variations, bottom). Video available at \url{https://tinyurl.com/ybfakpxf}.}
\vspace{-19pt}
\label{fig:teaser}
\end{figure}

As a step in this direction, we consider simple agents characterized by a single movement parameter that takes discrete ordinal values. 
Note that a variety of creative visualizations can be parameterized by a single value, including an agent moving along a $1$D grid, a pulsating disc, deforming geometric shapes, or a humanoid in a variety of sequential poses. 

In this work, we focus on designing interesting choreographies by combining the best of what humans are naturally good at -- heuristics of `good' dance that an audience might find appealing -- and what machines are good at -- optimizing well-defined objective functions.
Our intuition is that in order for a dance to go well with the music, the
overall spatio-temporal movement pattern should match the overall
structure of music.
That is, if the music is similar at two points in time, we would want the corresponding movements to be similar as well (Fig.~\ref{fig:teaser}).
We translate this intuition to an objective our agents optimize. 
Note that this is a flexible objective; it does not put constraints on the specific movements allowed. So there are multiple ways to dance to a music such that movements at points in time are similar when the music is similar, leaving room for discovery of novel dances.
We experiment with $25$ music clips from 13 diverse genres.
Our studies show that human subjects find our dances to be more creative compared to meaningful baselines.

%% file: sections/related_work.tex
\section{Related work}
\label{sec:related_work}

\textbf{Music representation.}
\cite{infantino2016robodanza,augello2017creative} use beat timings and loudness  as music features. 
We use Mel-Frequency Cepstral Coefficients (MFCCs) that capture fine-grained musical information.
\cite{yalta2019weakly} use the power spectrum (FFT) to represent music. 
MFCC features better match the exponential manner in which humans perceive pitch, while FFT has a linear resolution.

\textbf{Expert supervision.}
Hidden Markov Models have been used to choose suitable
movements for a humanoid robot to dance to a musical rhythm ~\cite{manfre2017learning}. 
\cite{lee2019dancing2music,lee2018listen,zhuang2020music2dance}
trained stick figures to dance by mapping music to human dance poses using neural networks. \cite{pettee2019beyond} trains models on human movement to generate novel choreography.
These works rely on data from 
dancers. 
In contrast, our approach does not require any expert supervision.

\textbf{Dance evaluation.}
\cite{tang2018dance} evaluate generated dance by asking users whether it matches the ``ground truth'' dance. This does not allow for creative variations in the dance. 
\cite{lee2019dancing2music,zhuang2020music2dance} evaluate their dances by asking subjects to compare a pair of dances based on beat rate, realism (independent of music), etc. 
Our evaluation focuses on whether human subjects find our generated dances to be creative and inspiring.

%% file: sections/dataset.tex
\section{Dataset}
\label{sec:dataset}
For most of our experiments, we created a dataset by sampling ${\sim}10$-second snippets from $22$ songs
for a total of $25$ snippets. We also show qualitative results for longer snippets towards the end of the paper.
To demonstrate the generality of our approach, we tried to ensure our dataset is as diverse as possible: our songs are sampled from 
1) $13$ different genres: Acapella, African, American Pop, Bollywood, Chinese, Indian-classical, Instrumental, Jazz, Latin, Non-lyrical, Offbeat, Rap, Rock 'n Roll, and have significant variance in 
2) number of beats: from complicated beats of Indian-classical dance of Bharatnatyam to highly
rhythmic Latin Salsa
3) tempo: from slow, soothing Sitar music to more upbeat Western Pop music
4) complexity (in number and type of instruments): from African folk music to Chinese classical
5) and lyrics (with and without).

%% file: sections/approach.tex
\section{Approach}
\label{sec:approach}

Our approach has four components --
the music representation,
the movement or dance representation (to be aligned with the music),
an alignment score,
and our greedy search procedure used to optimize this alignment score.

\begin{figure}[t!]
\begin{subfigure}{0.242\columnwidth}
  \includegraphics[width=0.99\columnwidth]{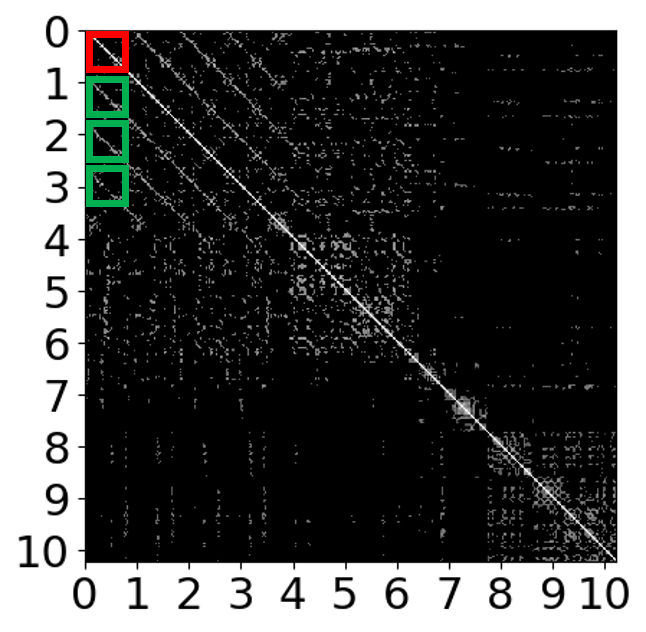}
\end{subfigure}
\begin{subfigure}{0.242\columnwidth}
  \includegraphics[width=0.99\columnwidth]{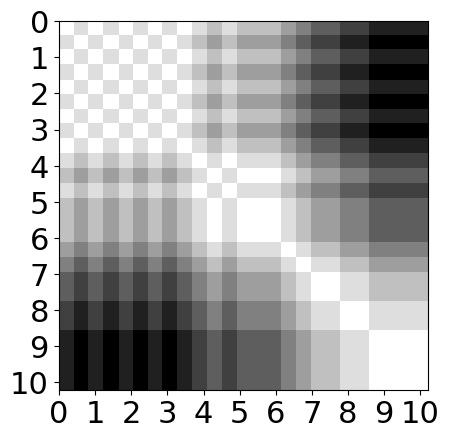}
\end{subfigure}
\begin{subfigure}{0.242\columnwidth}
  \includegraphics[width=0.99\columnwidth]{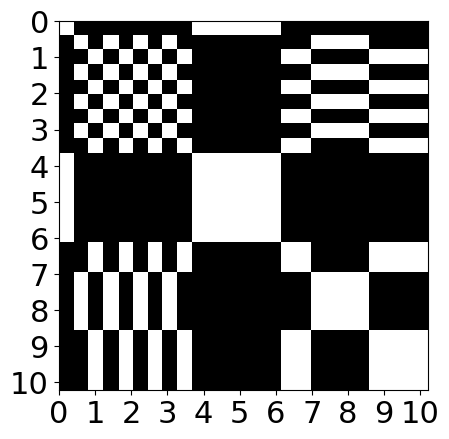}
\end{subfigure}
\begin{subfigure}{0.242\columnwidth}
  \includegraphics[width=0.99\columnwidth]{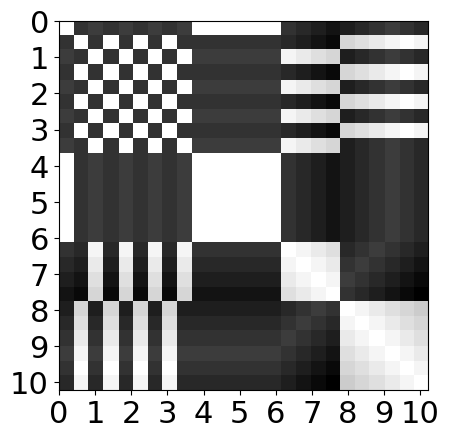}
\end{subfigure}
\caption{Music representation (left) along with  three dance representations for a well-aligned dance: state based (\state), action based (\action), state and action based (\stateplusaction).}
\vspace{-15pt}
\label{fig:music_dance}
\end{figure}

\noindent \textbf{Music representation.}
We extract Mel-Frequency Cepstral Coefficients (MFCCs) for each song. 
MFCCs are amplitudes of the power spectrum of the audio signal in Mel-frequency domain. 
Our implementation uses the Librosa library~\cite{mcfee2015librosa}.
We use a sampling rate of $22050$ and hop length of $512$.
Our music representation is a self-similarity matrix of the MFCC features, wherein
$\text{music}[i, j] = \text{exp}\left(-||\text{mfcc}_i-\text{mfcc}_j||_2\right)$
measures how similar frames $i$ and $j$ are.
This representation has been previously shown to
effectively encode structure~\cite{foote2003media} that is
useful for music retrieval applications.

Fig.~\ref{fig:music_dance} shows this music matrix for the song available at \url{https://tinyurl.com/yaurtk57}. 
A reference segment (shown in red, spanning $0.0$s to $0.8$s)
repeats several times later in the song (shown in green). 
Our music representation captures this repeating structure well.

\noindent \textbf{Dance representation.}
Our agent is parameterized with an ordinal movement parameter
$k$ that takes one of $1, ..., K$ discrete `states' at each step in a sequence of $N$ `actions'. 
The agent always begins in the middle $\sim\frac{K}{2}$.
At each step, the agent can take one of three actions: stay at the current state $k$, or move to adjacent states ($k-1$ or $k+1$) without going out of bounds.
We explore three ways to represent a dance.

1. State-based (\state). 
Similar to music, we define our dance matrix
$\text{dance}_{\text{state}}[i, j]$ as similarity in the agent's state at time $i$ and $j$: 
distance between the two states normalized by $(K-1)$, subtracted from 1. 
Similarity is 0 when the two states are the farthest possible, and 1 when they are the same. 

2. Action-based (\action). $\text{dance}_{\text{action}}[i, j]$ is $1$ when the agent takes the same action at times $i$ and $j$, and $0$ otherwise.

3. State $+$ action-based (\stateplusaction). 
As a combination, 
$\text{dance}_{\text{state+action}}$ is the average of $\text{dance}_{\text{state}}$ and $\text{dance}_{\text{action}}$. 

Reasoning about tuples of states and actions (as opposed to singletons at $i$ and $j$) is future work.

\noindent \textbf{Objective function: aligning music and dance.}
We use Pearson correlation between vectorized music and dance matrices as the objective function our agent optimizes to search for `good' dances.
Pearson correlation measures the strength of linear association between the two representations, and is high when the two matrices are aligned (leading to well-synchronized dance) and low if unsynchronized.

For an $M \times M$ music matrix and $N \times N$ dance matrix (where $N =$ no. of actions), we upsample the dance matrix to $M \times M$ via nearest neighbor interpolation
and then compute Pearson correlation. 
That is, each step in the dance corresponds to a temporal window in the input music.

In light of this objective, we can now intuitively understand
our dance representations.

State-based (\state):
Since this is based on distance between states, the agent is encouraged to position itself so that it revisits similar states when similar music sequences repeat. 
Note that this could be restrictive in the actions the agent can take or hard to optimize as it requires planning actions in advance to land near where it was when the music repeats.

Action-based (\action):
Since this is based on matching actions, the agent is encouraged to take actions such that it takes the same actions
when similar music sequences repeat. This has a natural analogy to human dancers who often repeat moves when the music repeats.
Intuitively, this is less restrictive than \state~because unlike states, actions are independent and not bound by transition constraints; recall that the agent can only move to adjacent states from a state (or stay).

\begin{figure}[t!]
\centering
\begin{subfigure}{0.31\columnwidth}
  \includegraphics[width=0.99\columnwidth]{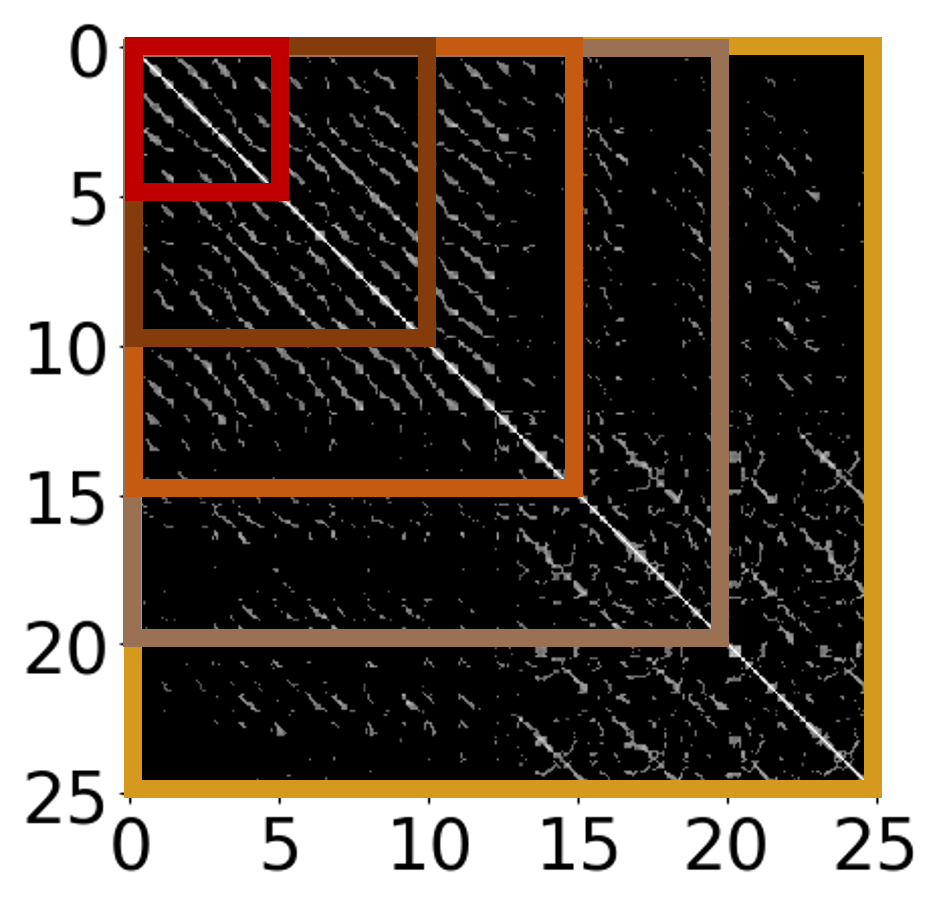}
\end{subfigure}
\hspace{20pt}
\begin{subfigure}{0.31\columnwidth}
  \includegraphics[width=0.99\columnwidth]{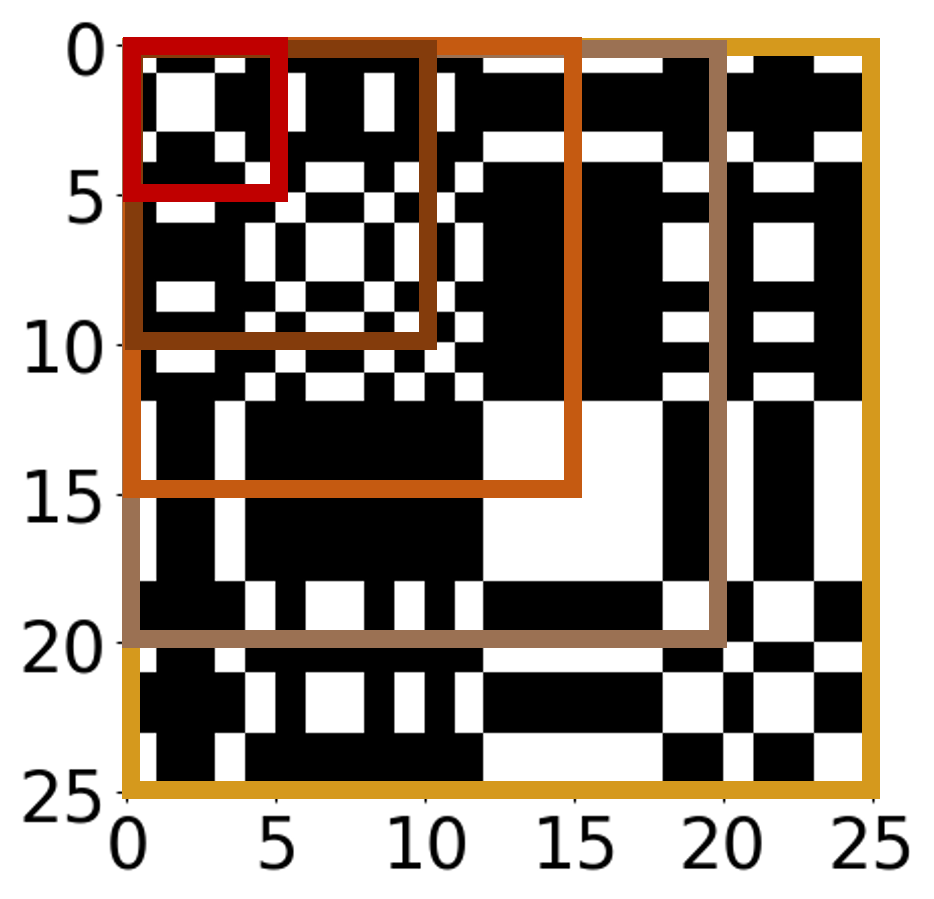}
\end{subfigure}
\caption{Our search procedure sequentially searches for dance sequences that result in high alignment between corresponding music (left) and dance (right) matrices. Sequential ordering shown as red to lighter shades of orange.}
\vspace{-15pt}
\label{fig:approach}
\end{figure}

\noindent \textbf{Search procedure.}
We use Beam Search with a single beam to find the best dance sequence given the music and dance matrices, as scored by the Pearson correlation objective described earlier. We use chunks of 5 dance steps as one node in the beam. The node can take one of $3^5$ values (3 action choices at each step).
Specifically, we start with the first $5$ steps and the corresponding music matrix (red boxes in Fig.~\ref{fig:approach}).
We compute Pearson correlation with all $3^5$ dance matrices, and return the best sequence for these $5$ steps.
Next, we set the first $5$ steps to the best sequence, and search over all combinations of the next $5$, \textit{i.e.}, $3^5$ sequences, each of length $10$ now.
See orange boxes in Fig.~\ref{fig:approach}. 
This continues till a sequence of length $N$ has been found (i.e., the music ends). Our music and dance representations scale well with song length. 
Our search procedure scales linearly with number of steps in the dance. While its greedy nature allows the agent to dance $\sim$live with the music, it may result in worse synchronization for later parts of the song. It scales exponentially with number of actions, and we discuss approaches to overcome this in Future Work.

%% file: sections/evaluation.tex
\section{Baselines}
\label{sec:baselines}

Creativity is often thought of as a combination of novelty and quality~\cite{boden92}.
We hypothesize that dances where an agent moves predictably (\textit{i.e.} not surprising/novel) or that are not synchronized with the music (\textit{i.e.} low quality) will be deemed less creative.
This motivates our baselines:

1. Synced, sequential (\syncseq). The agent moves sequentially from one extreme of the state space to the other till the music ends. It only moves when there is a beat in the music.
The beat information is extracted using the implementation of \cite{krebs2015efficient} from the Madmom library.
This baseline is synced, yet predictable and uninteresting.

2. Un-synced, sequential (\unsyncseq).
The agent also moves sequentially, but ignores the beats and moves at every step.
This baseline is unsynced and predictable.

3. Synced, random (\syncran).
The agent takes a random action from its allowed actions at every beat, and stays put otherwise.
This baseline is synced and quite unpredictable, so we expect
this to be more interesting than \syncseq~and \unsyncseq.

4. Un-synced, random (\unsyncran).
The agent takes a random action from its allowed actions independent of when the beat occurs.
This baseline is unsynced and unpredictable.

\section{Evaluation via human studies}
\label{sec:evaluation}

We compare our approach using the $3$ dance representations against the $4$ baselines for $25$ song snippets and values of $N \in \{25, 50, 100\}$ (no. of steps in the dance).
We set the number of states an agent can be in to $K = 20$.
For this experiment, we visualize the agent
as a dot, with state indicating a location on a $1$-D grid.
We first compare approaches with the same $N$, and then for the best approach, compare different $N$s.
We then compare our best approach to the strongest baseline using other
visualizations to evaluate the role visualization plays in perception of dance creativity.
For each of these settings, we show subjects on Amazon Mechanical Turk (AMT) a pair of dances and ask them: Which dance
    (1) goes better with the music?
    (2) is more surprising / unpredictable?
    (3) is more creative?
    (4) is more inspiring?
Subjects can pick one of the two dances or rate them equally.

\textbf{Dance representation.}
The $7$ approaches amount to $7 \choose 2$ = $21$ pairs of dances per song per $N$.
We showed each pair (for the same song and $N$)
to $5$ subjects on AMT. For the $25$ songs and $N \in \{25, 50, 100\}$, this results in a total of $7875$ comparisons.
$210$ unique subjects participated in this study.

\begin{figure}[t!]
	\centering
	\includegraphics[width=0.99\columnwidth]{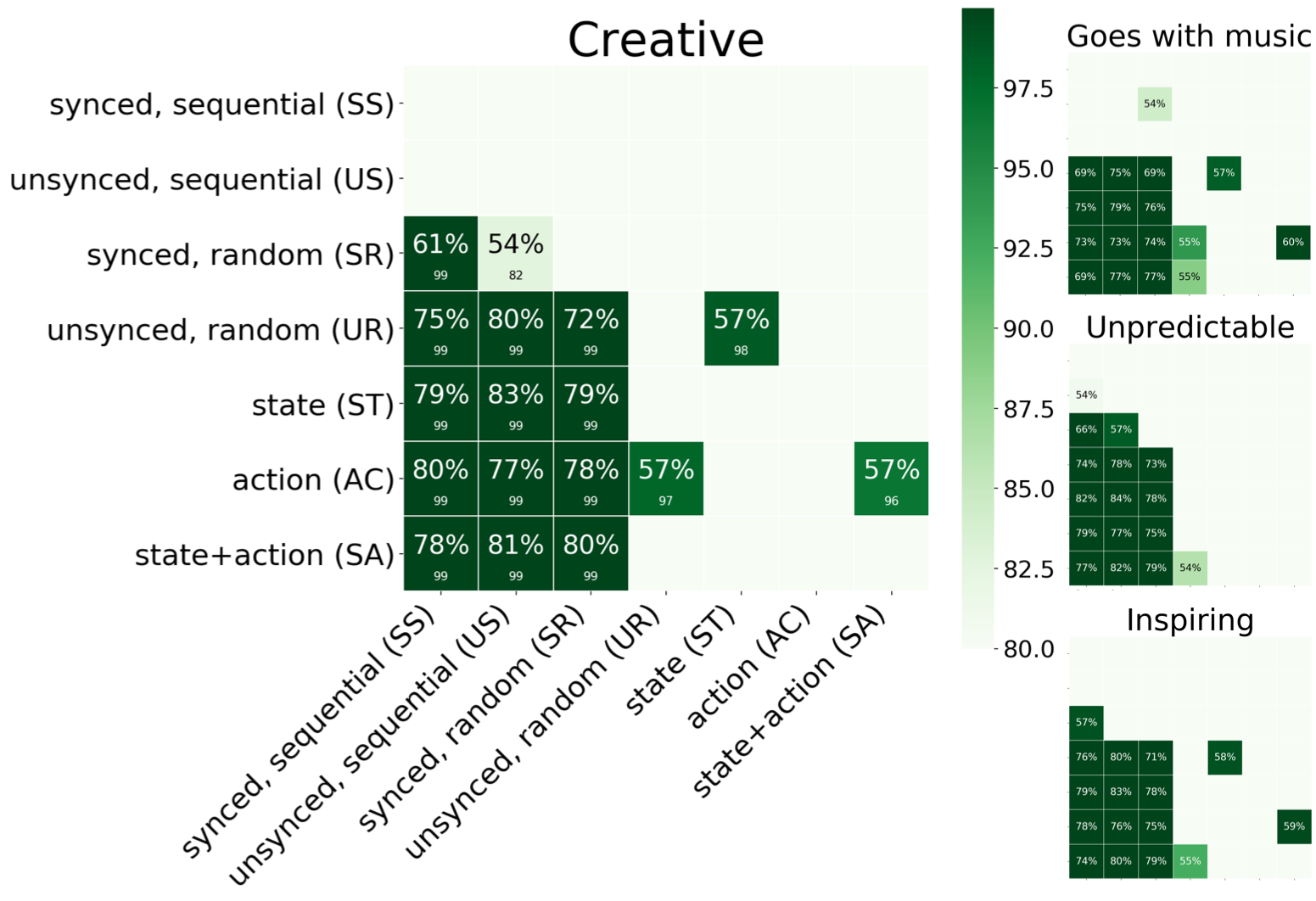}
    \caption{Evaluation via human studies of dances on 4 metrics --
    a) creativity, b) synchronization with music, c) unpredictability, and d) inspiration.
    Table cells show win rate of approach in row against approach in column.
    Shade of green and subscript shows statistical confidence (only for $>80\%$).}
	\label{fig:compare}
	\vspace{-15pt}
\end{figure}

See Fig.~\ref{fig:compare}.
Table cells show win rate of approach in row against approach in column.
Subscripts and green shades show statistical confidence levels (shown only for $> 80\%$).
For example, dances from \syncran~are found to be more creative than those from~\syncseq~$61\%$ of the times. That is, at our sample size, \syncran~ is more creative than \syncseq~ with $99\%$ confidence.
Among baselines (rows $1$ to $4$), humans found random variants (\syncran, \unsyncran)
to be more unpredictable (as expected), and \unsyncran~to be more creative,
better synchronized to music, and more inspiring than sequential variants (\syncseq, \unsyncseq).
\unsyncran~is the best-performing baseline across metrics.
We hypothesize that \unsyncran~performs better than \syncran~because the latter only moves with beats. Comments from subjects indicate that they prefer agents that also move with other features of the music.
All our proposed approaches perform better than~\syncseq, \unsyncseq, \syncran~baselines across metrics.
\action~is rated comparable to \state~and \stateplusaction~in terms of (un)predictability. But more creative, synchronized with music, and inspiring.
This may be because as discussed earlier, state-based synchronization is harder to achieve. Moreover, repetition in actions for repeating music is perhaps more common among dancers than repetition in states (location). Finally, our best approach \action~is rated as more creative than the strongest baseline \unsyncran.

\textbf{Number of steps.}
With a higher number of steps, the agent can sync to the music with higher precision.
However more steps would add more ``jumpiness'' to the dance, which may not be desirable.
We evaluate our best approach (\action~) for $N \in \{25, 50, 100\}$.
This gives us $3 \choose 2$ = 6 pairs of dances per song.
We showed each pair for each of the 25 songs to 5 AMT subjects; 375 pairwise comparisons from 22 unique subjects.
Subjects find dances with 100 steps to be more creative than 50 and 25 steps at 99\% statistical confidence, with 100 steps preferred 69\% and 73\% of the times respectively.

\textbf{Effect of visualizations.}
Finally, we analyze how choice of visualization affects perception of dance creativity. We compare our best approach (\action)~with the strongest baseline (\unsyncran)~for 6 different visualizations including a pulsating disc, a stick figure, and collections of deforming geometric shapes. Including the dot on a $1$-D grid from earlier, we have 7 pairs of dances for 25 songs and 5 evaluators; 875 comparisons from 59 unique subjects.
Preference for our approach for creativity ranges from 48\% to 61\% across visualizations, with 2 visualizations at $<$50\%. Preference on only 3 of the 7 visualizations is significant at $>$95\% confidence; all favor our approach. Interestingly, one of these visualizations corresponds to a human stick figure. Perhaps nuances in dance are more easily perceived with human-like visualizations.

Example dances of our best approach (\action) for different songs, number of steps, visualizations, and song durations can be found at \url{https://tinyurl.com/ycoz6az8}.

%% file: sections/discussion.tex
\section{Discussion}
\label{sec:discussion}
Our preliminary study with a simple agent gives promising indications that subjects find dances discovered using our flexible, intuitive heuristic to be creative. The next step is to train more complex agents to dance. Our search-based approach will not scale well with larger action spaces. We plan to use machine learning approaches to optimize for the music-dance alignment, so that given a new song at test time, an aligned dance sequence an be produced without an explicit search. Rather than supervised learning approaches described in Related Work which require annotated data, we will explore Reinforcement Learning (RL) using our objective function as a reward. This retains the the possibility of discovering novel dances, which is central to creativity.